\documentclass{article}

\usepackage{arxiv}

\usepackage[utf8]{inputenc} 
\usepackage[T1]{fontenc}    
\usepackage{hyperref}       
\usepackage{url}            
\usepackage{booktabs}       
\usepackage{amsfonts}       
\usepackage{nicefrac}       
\usepackage{microtype}      
\usepackage{lipsum}
\usepackage{subfigure}
\usepackage{graphicx}
\usepackage{amsmath}
\usepackage{cleveref}
\graphicspath{{Figures/}}
\usepackage{xcolor}
\usepackage{mdframed}

\newcommand{\mbf}{\mathbf}
\newcommand{\mbb}{\mathbb}
\newcommand{\mcl}{\mathcal}
\newcommand{\f}{\frac}
\newcommand{\bs}{\boldsymbol}
\newcommand{\RNum}[1]{\uppercase\expandafter{\romannumeral #1\relax}}

\newcommand{\T}{\textnormal}
\newcommand{\x}{\mathbf{x}}

\newcommand{\btht}{\boldsymbol{\theta}}
\newcommand{\Tht}{\Theta}
\newcommand{\bTht}{\boldsymbol{\Theta}}

\title{Data-Driven Wind Turbine Wake Modeling via Probabilistic Machine Learning}

\author{
  S. Ashwin Renganathan\thanks{Argonne National Laboratory} \\
  Mathematics \& Computer Science \\
  \texttt{srenganathan@anl.gov} \\
  \And
    Romit Maulik \thanks{Argonne National Laboratory} \\
  Mathematics \& Computer Science \\
  \texttt{rmaulik@anl.gov} \\
  \And
Stefano Letizia \\
  Department of Mechanical Engineering\\
  The University of Texas at Dallas,\\
  Dallas, TX \\
  \texttt{stefano.letizia@utdallas.edu} \\
    \And
 Giacomo Valerio Iungo \\
  Department of Mechanical Engineering\\
  The University of Texas at Dallas,\\
  Dallas, TX \\
  \texttt{valerio.iungo@utdallas.edu} \\
}

\begin{document}
\maketitle

\begin{abstract}
Wind farm design primarily depends on the variability of the wind turbine wake flows to the atmospheric wind conditions, and the interaction between wakes. Physics-based models that capture the wake flow-field with high-fidelity are computationally very expensive to perform layout optimization of wind farms, and, thus, data-driven reduced order models can represent an efficient alternative for simulating wind farms. In this work, we use real-world light detection and ranging (LiDAR) measurements of wind-turbine wakes to construct predictive surrogate models using machine learning. Specifically, we first demonstrate the use of deep autoencoders to find a low-dimensional \emph{latent} space that gives a computationally tractable approximation of the wake LiDAR measurements. Then, we learn the mapping between the parameter space and the (latent space) wake flow-fields using a deep neural network. Additionally, we also demonstrate the use of a probabilistic machine learning technique, namely, Gaussian process modeling, to learn the parameter-space-latent-space mapping in addition to the epistemic and aleatoric uncertainty in the data. Finally, to cope with training large datasets, we demonstrate the use of variational Gaussian process models that provide a tractable alternative to the conventional Gaussian process models for large datasets. Furthermore, we introduce the use of active learning to adaptively build and improve a conventional Gaussian process model predictive capability. Overall, we find that our approach provides accurate approximations of the wind-turbine wake flow field that can be queried at an orders-of-magnitude cheaper cost than those generated with high-fidelity physics-based simulations.
\end{abstract}

\section{Introduction}
\label{s:introduction}

Understanding and modeling wind farm flows still represent major challenges for wind farm designers and operators. Important aspects, such as the interaction of the atmospheric boundary layer with wind turbines \cite{SanzRodrigo2017,Veers2019} and prediction of the wake morphology and their superposition \cite{Porte-agel2019}, are far from being fully understood, in spite of having been the object of numerous insightful scientific studies.

The term turbine wake refers to the low momentum and highly turbulent region located downstream of an operating wind turbine, which is a direct consequence of the extraction of kinetic energy from the incoming wind field. Significant power losses \cite{Barthelmie2010,El-Asha2017,Sebastiani2020} and enhanced fatigue loads \cite{Churchfield2012,Conti2020} were documented for turbines impinged by upstream wakes. The study of turbine wakes through numerical simulations is encumbered with difficulties due to the high Reynolds number flow (which entails a great span of length and time scales involved) \cite{Porte-agel2019}, the unsteadiness of the inflow conditions \cite{Iungo2014,Letizia2021}, and the relevant role of atmospheric stability \cite{zhan2020lidar}. Recently, large-eddy simulations (LES) have become a well-established tool for the simulation of wind farm flows \cite{Mehta2014,Breton2017,Santoni2020}. However, their computational costs are still prohibitive for large scale tasks, such as for layout optimization and real-time performance diagnostics.

Reynolds-averaged Navier-Stokes \cite{Sanderse2011,iungo2018parabolic} and engineering wake models (e.g. \cite{Jensen1983,Frandsen2006,Bastankhah2014}) provide faster and simpler alternatives to LES, since they do not explicitly solve the unsteady small-scale turbulent eddies. However, while the former are oftentimes affected by inaccuracy due to the turbulence closure \cite{Sanderse2011}, the latter require a thorough calibration to achieve a satisfactory agreement with experimental data \cite{Zhan2020}. 

The above-mentioned challenges have spurred the interest of wind energy scientists in the experimental characterization of the wind farm flow. In particular, the improvements of remote sensing instruments, such as wind Light Detection and Ranging (LiDAR), have promoted the proliferation of field experimental campaign investigating the wakes of utility-scale wind turbines (e.g., see \cite{Iungo2014,Machefaux2016,Zhan2020}). These studies have highlighted the great complexity and sensitivity to environmental conditions on the characteristics and downstream evolution of wind turbine wakes.

Following up on past work~\cite{maulik2021cluster}, in this work, LiDAR measurements of individual wakes generated by utility-scale wind turbines under broad ranges of atmospheric and wind conditions are leveraged to develop data-driven machine learning models to enable accurate predictions of the wake velocity field for prescribed wind/atmospheric conditions, while requiring computational costs as low as for empirical wake engineering models. Specifically, we explore the application of deep neural networks (DNN) for efficient data reduction via autoencoders, as well as to build predictive models via multilayer perceptrons. Furthermore, we also explore the combination of the data reduction via the DNN and Gaussian process (GP) models to develop predictive probabilistic models of the wake flow. This way, we show how probabilistic machine learning can be leveraged to learn predictive wake models from noisy and incomplete measurement data. To summarize, the contributions of this article are:

\begin{enumerate}
    \item A novel convolutional autoencoder framework to obtain low-dimensional embeddings of wind LiDAR measurements for wind-turbine wakes. In this manner, we develop a compressed---and hence tractable---representation of the high-dimensional LiDAR data.
    \item{\label{contrib:DNN}} A DNN to learn the mapping between the input parameter space and the latent space derived from the convolutional autoencoder. Therefore, we learn the mapping between the input parameters and the spatially distributed wake measurements, via two levels of DNNs--one for data reduction and one for prediction. This serves as a cheap-to-evaluate surrogate model to predict wind turbine wake velocity fields.
    \item Additionally, similar to \cref{contrib:DNN}, we propose the use of probabilistic machine learning model via GP regression to learn latent-space to wake flow field mapping, which can simultaneously learn the noise in the data. To address the tractability issues associated with GP regression with \emph{big} data, we also investigate the use of \emph{variational} and active learned GP regression.
\end{enumerate}

\begin{figure}[htb!]
    \centering
    \includegraphics[width=1\linewidth]{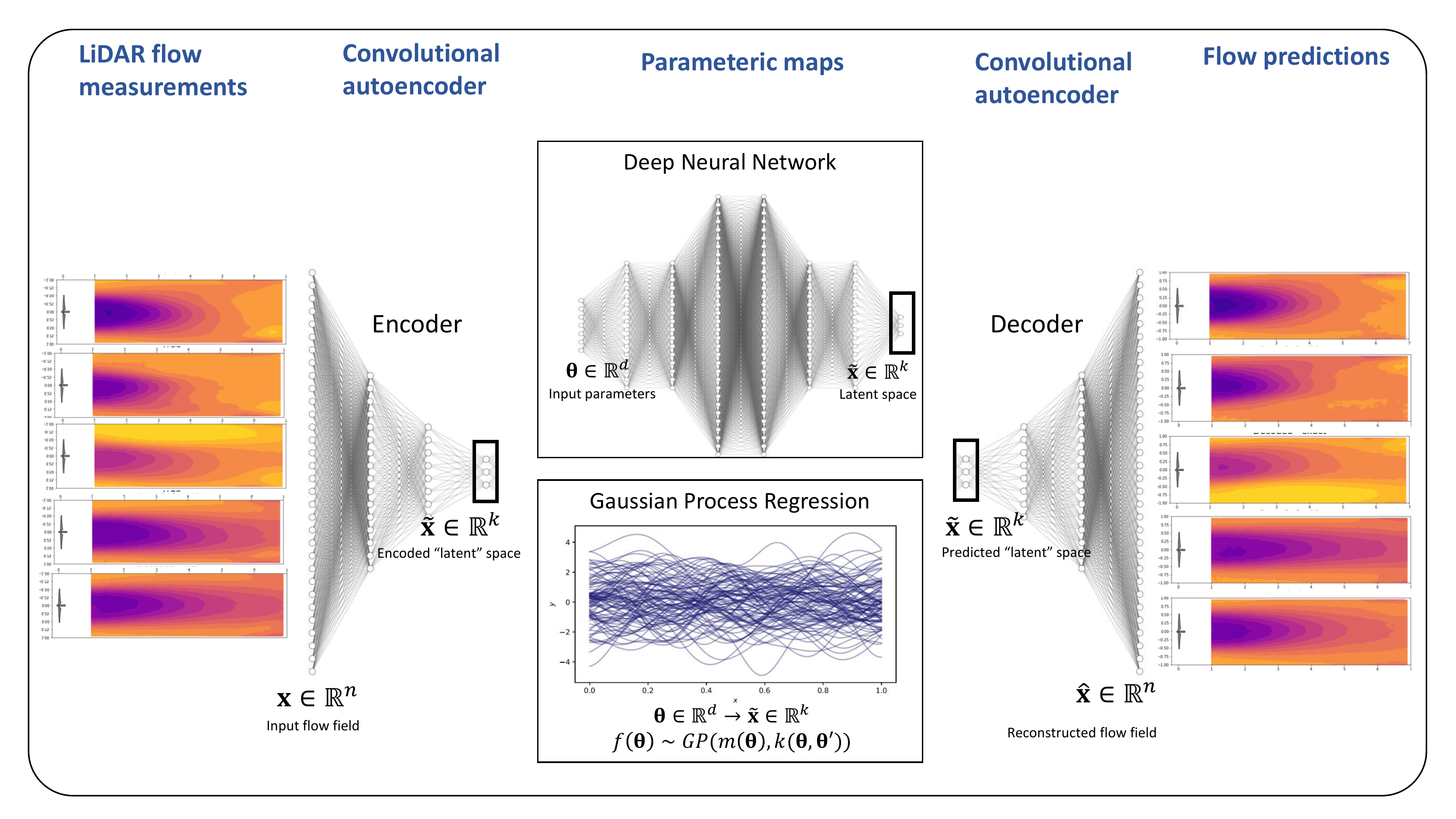}
    \caption{Summary of the overall methodology.}
    \label{fig:method}
\end{figure}

Overall, this work brings together state-of-the-art data-driven machine learning and the classic field of wind energy to build cheap and reliable surrogate models that can be leveraged to perform exploratory studies. \cref{fig:method} provides a high-level summary of the work.

The remainder of this article is organized as follows. In \cref{s:data_collection}, we present the details of our experimental setup and data collection via LiDAR. In sections \cref{s:deep_learning} and \cref{s:gpr} we present the theoretical details of our machine learning methods. We present the results and associated discussion in \cref{s:results}, followed by concluding remarks and an outlook on future work in \cref{s:conclusions}.

\section{Experimental data and collection methods}
\label{s:data_collection}

The wind LiDAR measurements used in this work were collected in the period from August 2015 to March 2017 at a wind farm located in North Texas~\footnote{The original LiDAR data used in this work are available upon reasonable request from the fourth author, who may be contacted at \texttt{valerio.iungo@utdallas.edu.}}. The wind farm includes 25 identical wind turbines with a nameplate capacity of 2.3 MW. The rotor diameter is $d=127$ m and the height of the hub is 89 m above the ground. The local topographic map provided by \cite{USGS} with a resolution of 100 m shows that 95\% of the terrain within the farm has a slope lower than $3^\circ$, which allows to rule out effects due to the terrain on the flow. 

\begin{figure}[h!]
    \centering
    \includegraphics[width=1.4\linewidth]{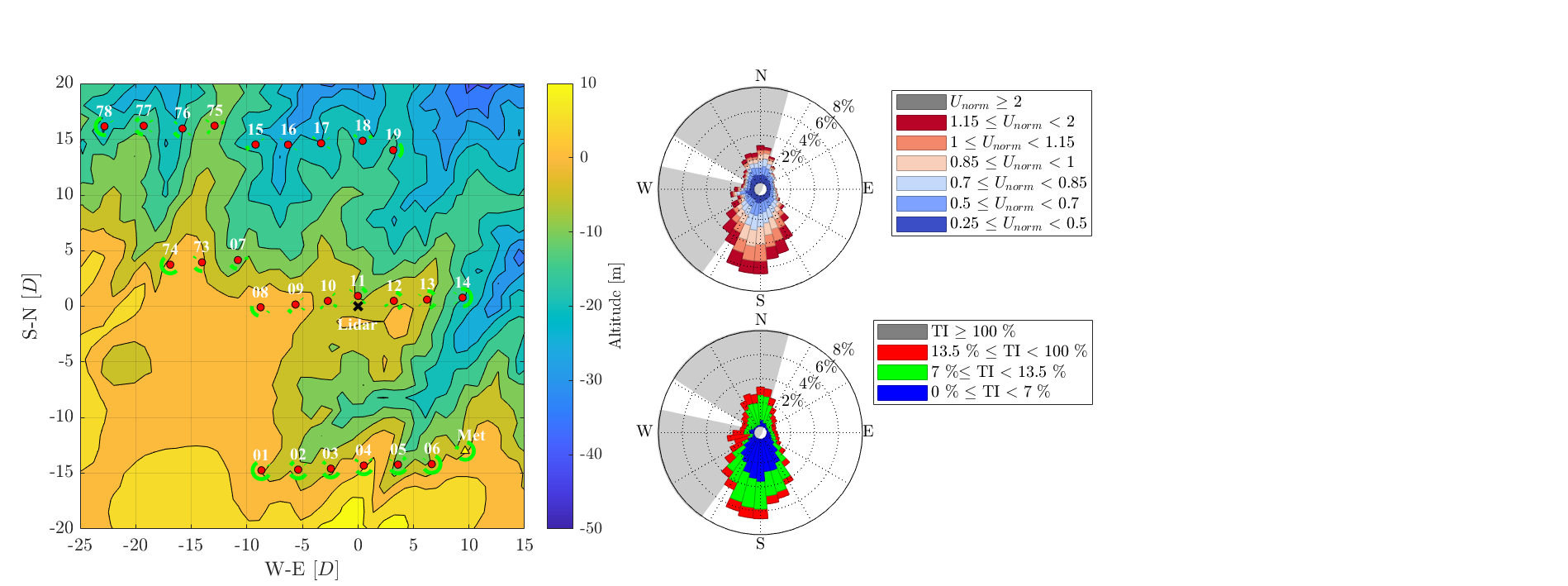}
    \caption{Map of the experimental site. On the left, is the topographic map and farm layout, where the green arcs indicate wind sectors where no relevant wake interactions are expected. On the right, it is the directional histogram of wind speed and turbulence intensity based on the meteorological data of the 80-meters tower at the "Met" location. $U_{norm}$ refers to the hub-height wind speed normalized by the rated wind speed of the turbines (11 m s$^{-1}$). Grey sectors are likely affected by wakes.}
    \label{fig:Pandandle_map}
\end{figure}

A WindCube 200S scanning pulsed D\"{o}ppler LiDAR was installed at turbine 11 (see \cref{fig:Pandandle_map}) and probed the wakes stemming from turbines 01 to 06 during the occurrence of southerly winds. Plan Position Indicator (PPI) scans were scheduled targeting the wake of the turbine with the best line-of-sight alignment with the LiDAR, which enables achieving an optimal spatio-temporal sampling resolution. 

Furthermore, meteorological and SCADA data were continuously collected for the whole duration of the campaign in the form of 10-minute mean and standard deviation of wind speed, wind direction, temperature, atmospheric pressure, active power, RPM, and blade pitch angle. The list of input parameters is presented in \cref{tab:input_parameters}; for a more detailed description of the experimental site and LiDAR scanning strategy please refer to \cite{El-Asha2017,zhan2020lidar}.

The wind rose of the site (\cref{fig:Pandandle_map}) indicates high likelihood of southerly winds and a negligible directional dependence of the turbulence intensity at hub-height, which confirms the findings of previous studies \cite{El-Asha2017,zhan2020lidar} on the prevalent thermally-driven cycle of the atmospheric boundary layer caused by the diurnal variation of the solar irradiance. 

The LiDAR data undergo a quality control which excludes all the points characterized by a carrier-to-noise ratio lower than $-25$ dB. Subsequently, the wake data are realigned with the wind direction, which is estimated as the 10-minute moving averaged wake direction. This also allows estimating the horizontal equivalent velocity as:
\begin{equation}\label{eq:Eq_velocity}
u_{eq}\sim \frac{u_\text{LOS}}{\cos(\theta-\theta_w)\cos\beta},
\end{equation}
where $u_\text{LOS}$ is the line-of-sight velocity measured by the LiDAR, $\theta_w$ is the wind direction, $\theta$ and $\beta$ are the azimuth and elevation angles of the LiDAR, respectively. The vertical variability of the incoming wind due to wind shear is corrected by normalizing the equivalent velocity by the vertical undisturbed velocity profile. After the outlined post-processing, 6,654 quality-control re-aligned and non-dimensional LiDAR scans are made available for the following analysis. For more technical details on this procedure, the interested reader shall refer to \cite{zhan2020lidar}.
\begin{table}[htb!]
    \centering
    \begin{tabular}{l|c|c}
        \hline
         Parameter & Range & Description  \\
         \hline
         \hline
         SCADA\_WS [m/s]& $[2.92, 15.22]$ & Mean Hub-height wind speed recorded by the SCADA\\
         MET\_WS\_80m [m/s]& $[3.9,15.22]$ & Mean Hub-height wind speed recorded by the met-tower\\
         SCADA\_TI [-]& $[0.04,0.36]$ & Hub-height wind turbulence intensity recorded by the SCADA\\
         MET\_BulkRichardson [-]& $[-0.01,  0.01]$ & Bulk Richardson number recorded by met-tower\\
         SCADA\_Power [kW]& $[58.8, 2423]$ & Mean power capture recorded by the SCADA\\
         SCADA\_RPM [RPM]& $[7.07, 16.95]$ & Mean rotor rotational speed recorded by the SCADA\\
         SCADA\_Pitch [deg]& $[-2,80]$ & Mean blade pitch angle recorded by the SCADA\\
         \hline
    \end{tabular}
    \caption{List of input parameters and their ranges.}
    \label{tab:input_parameters}
\end{table}

\section{Deep learning for parametric flow prediction}
\label{s:deep_learning}

In the following section, we introduce our deep neural network architectures for establishing a viable emulation strategy for data obtained from LiDAR measurements.

\subsection{Convolutional autoencoder}
\label{ss:conv_autoencoder}

Autoencoders are neural networks that learn a new representation of the input data, usually with lower dimensionality. The initial layers, called the \emph{encoder}, map the input $\x \in \mathbb{R}^n$ to a new representation $\tilde{\x} \in \mathbb{R}^k$ with $k \ll n$. The remaining layers, called the \emph{decoder}, map $\tilde{\x}$ back to $\mathbb{R}^n$ with the goal of reconstructing $\x$. The objective is to minimize the reconstruction error. Autoencoders are unsupervised; the data $\x$ is given, but the representation $\tilde{\x}$ must be learned.

More specifically, we use autoencoders that have convolutional layers. In a convolutional layer, instead of learning a matrix that connects all $m$ neurons of layer's input to all $n$ neurons of the layer's output, we learn a set of filters that are convolved with regions of the layer's input. Suppose a one-dimensional (1-d) convolutional layer has filters of length $L_f$, then, each of the layer's output neurons corresponding to a specific filter $\mathbf{f^i}$ is connected to a patch of $L_f$ of the layer's input neurons. In particular, a 1-d convolution of filter $\mathbf{f^i}$ and patch $\mathbf{p}$ is defined as $\mathbf{f^i} \ast \mathbf{p} = \sum_j f^i_j p_j$ (where $f^i_j$ corresponds to the stencil coefficient in the filter for index $j$). In other words, convolutional neural networks identify stencil values $f^i_j$ that obtain coherent translationally invariant features relevant to a particular function approximator. Then, for a typical 1-d convolutional layer, the layer's output neuron $y_{ij} = \varphi (\mathbf{f^i} \ast \mathbf{p_j} +B_{i})$ where $\varphi$ is an activation function, and $B_i$ are the entries of a bias term. As $\mathbf{j}$ increases, patches are shifted by stride $s$. For example, a 1-d convolutional layer with a filter $\mathbf{f^0}$ of length $m_f = 3$ and stride $s=1$ could be defined so that $y_{0j}$ involves the convolution of $\mathbf{f^0}$ and inputs $\mathbf{j}-1, \mathbf{j}$, and $\mathbf{j}+1$. To calculate the convolution, it is common to add zeros around the inputs to a layer, which is called \emph{zero padding}. In the decoder, we use deconvolutional layers to return to the original dimension. These layers upsample with nearest-neighbor interpolation.

Two-dimensional convolutions are defined similarly, but each filter and each patch are two-dimensional. A 2-d convolution sums over both dimensions, and patches are shifted both ways. For a typical 2-d convolutional layer, the output neuron $y_{hij} = \varphi (\mathbf{f^h} \ast \mathbf{p_{ij}} +B_{h})$. Input data can also have a ``channel'' dimension, such as Red/Green/Blue values for images. The convolutional operator sums over channel dimensions, but each patch contains all of the channels. The filters remain the same size as patches, so they can have different weights for different channels. It is common to follow a convolutional layer with a \emph{pooling} layer, which outputs a sub-sampled version of the input. In this paper, we specifically use max-pooling layers. Each output of a max-pooling layer is connected to a patch of the input, and it returns the maximum value in the patch. 

Autoencoders have recently become popular for the nonlinear dimensionality reduction of datasets extracted from several high dimensional systems. These have been motivated by the extraction of coherent structures that parameterize low-dimensional embeddings in manifolds \cite{fukami2020convolutional,murata2020nonlinear,vennemann2020dynamic}, and the utilization of these embeddings for efficient surrogate models of nonlinear dynamical systems \cite{gonzalez2018deep,maulik2021reduced,kim2020fast,wu2021reduced,cheng2020advanced,renganathan2018koopman,renganathan2020koopman,renganathan2020machine}. In this work, we utilize convolutional autoencoders to identify low-dimensional representations of experimentally collected data for building parameter-observation maps where the former are obtained through meteorological and wind turbine data and the latter are LiDAR measurements collected in the wake generated by wind turbines.

\subsection{Multilayered perceptron (MLP)}
\label{ss:mlp}


One technique to obtain a mapping from the meteorological and turbine datasers and the latent space embeddings of the convolutional autoencoder is through the use of a multilayered perceptron (MLP) architecture, which is a subclass of feedforward artificial neural network. A general MLP consists of several neurons arranged in multiple layers. These layers consist of one input and one output layer along with several hidden layers. Each layer (with the exception of an input layer) represents a linear operation followed by a nonlinear activation that allows for great flexibility in representing complicated nonlinear mappings. This may be expressed as
\begin{align}
    \mathcal{L}^{l}\left(\boldsymbol{t}^{l-1}\right):=\boldsymbol{w}^{l} \boldsymbol{t}^{l-1}+\boldsymbol{b}^{l},
\end{align}
where $\boldsymbol{t}^{l-1}$ is the output of the previous layer, and $\boldsymbol{w}^l,\boldsymbol{b}^l$ are the weights and biases associated with that layer. The output $\boldsymbol{t}^l$, for each layer may then be transformed by a nonlinear activation, such as rectified linear activation:
\begin{align}
    \eta(a) = \text{ReLU} (a) = \max(a,0).
\end{align}
For our experiments, the inputs $\boldsymbol{t}^0$ are in $\mathbb{R}^d$ (i.e., $d$ is the number of inputs) and the outputs $\boldsymbol{t}^K$ are in $\mathbb{R}^k$ (i.e., $k$ is the number of outputs). The final map is given by
\begin{equation}
    \begin{split}
        F: \mathbb{R}^{d} \mapsto \mathbb{R}^{k} &, \quad \boldsymbol{t}^0 \mapsto \boldsymbol{t}^K = F(\boldsymbol{t}^0 ;(\boldsymbol{w}, \boldsymbol{b})), \\
        \text{where} \\
        F(\boldsymbol{t}^0 ; \boldsymbol{w},\boldsymbol{b})=&\eta^{K}\left(\mathcal{L}^{K} \circ \eta^{K-1}_{\text{act}} \circ \mathcal{L}^{K-1} \circ \ldots \circ \eta^{1}_{\text{act}} \circ \mathcal{L}^{1}\right)(\boldsymbol{t}^0)
    \end{split}
\end{equation}

is a complete representation of the neural network and where $\mathbf{w}$ and $\mathbf{b}$ are a collection of all the weights and the biases of the neural network. These weights and biases, lumped together as $\phi = \{\mathbf{w}, \mathbf{b} \}$, are trainable parameters of our map, which can be optimized by examples obtained from a training set. The supervised learning framework requires for this set to have examples of inputs in $\mathbb{R}^{N_{ip}}$ and their corresponding outputs $\mathbb{R}^{N_{op}}$. This is coupled with a cost function $\mathcal{C}$, which is a measure of the error of the prediction of the network and the ground truth. Our cost function is given by
\begin{align}
    \mathbb{C} = \frac{1}{|T|} \sum_{(\bs{\theta}, \tilde{\x}) \in T}\left\|\tilde{\x} - F(\bs{\theta};\phi)\right\|^{2}
\end{align}
with $|T|$ indicates the cardinality of the training data set given by
\[    T=\left\{\left(\bs{\theta}_i, \tilde{\x}_i\right): \tilde{\x}_i=f\left(\bs{\theta}_i\right)\right\}.
\]
and where $f\left(\bs{\theta}_i\right)$ are examples of the true targets obtained from the compressed training data using the autoencoder introduced in the previous sections. Gradients of this cost function can then be used in an optimization framework to obtain the best weights and biases, given the training data. Finally, the trained MLP may be used for uniformly approximating any continuous function on compact domains \cite{cybenko1989approximation,barron1993universal}, provided $\eta(x)$ is not polynomial in nature.

\section{Gaussian process regression}
\label{s:gpr}

We now review the preliminaries of GP models. Our primary interest in the use of GP models stems from its promise of offering enhanced data efficiency in emulation compared to DNNs~\cite{renganathan2021enhanced} as well as in sequential decision-making~\cite{renganathan2021lookahead}. Although not pursued in this work, GPs offer greater potential in emulating complex functions when combined with DNNs; e.g., see ~\cite{rajaram2020deep,rajaram2021empirical}. GP models provide a probabilistic approximation to an unknown function $f(\btht)$. Specifically, $f(\btht)$ is assumed to take the form of a GP, where each realization (or sample path) is a function. This \emph{prior} assumption on the function can then be combined with the probability of actual observations conditional on the prior (a.k.a, the \emph{likelihood}), using Bayes' rule~\cite{gelman1995bayesian}. In this section, we provide a brief overview of the theory behind GP models and highlight the difference between exact and approximate inference, the latter finding applications in the presence of large datasets.

\subsection{Exact Gaussian process regression}
\label{ss:exact_gp}
We begin by placing a GP prior assumption on the unknown function, that is, $f(\btht) \sim \mcl{GP}(\mu(\btht), k(\btht, \btht'))$, where $\mu(\btht)$ is a mean function and $k(\btht,\btht')$ is a covariance function (or \emph{kernel}), and $\bTht \in \mcl{T}$. We assume that we have noisy observations of $f(\btht)$ of the form
\[y_i = f(\btht) + \epsilon_i, \quad i=1,\ldots,n\]
where $\epsilon_i$ represents the observation noise. We assume $\epsilon_i$ to take an independent and identically distributed normal distribution with zero mean and a variance of $\sigma_\epsilon^2$, that is, $\epsilon_i \sim \mcl{N}(0, \sigma_\epsilon^2)$. Furthermore, we assume a Gaussian likelihood that defines the probability of observing the data given the GP assumption.


Let $\mbf{y}_n = [y_1,\ldots,y_n]^\top$ denote the vector of noisy observations of $f$ at $\bTht = [\btht_1,\ldots,\btht_n ]^\top$, and $\mcl{D}_n = \{(\btht_i, y_i), ~i=1,\ldots,n\}$, then, applying Bayes' rule, the posterior distribution \cite{williams2006gaussian} is given by
\begin{equation}
    \begin{split}
        f(\btht)|\mcl{D}_n \sim& \mcl{GP}(\mu_n(\btht), \sigma_n^2(\btht)),\\
        \T{where~}\mu_n(\btht) =& \mbf{k}^\top \mbf{K}^{-1} (\mbf{y}_n - \mu(\bTht)) \\
        \sigma_n^2(\btht) =& k(\btht, \btht) - \mbf{k}^\top \mbf{K}^{-1} \mbf{k}.
    \end{split}
    \label{e:posterior}
\end{equation}
In \eqref{e:posterior}, $\mu_n$ and $\sigma_n^2$ are the mean and variance of the posterior distribution, $\mbf{K} \subset \mcl{K}$ is the $n\times n$ covariance matrix with $\mbf{K}_{ij} = k(\btht_i, \btht_j), \forall \btht_i, \btht_j \in \bTht$ and $\mcl{K}$ being the cone of symmetric positive definite matrices,
$\mu({\bTht}) = [\mu(\btht_1), \ldots, \mu(\btht_n)]^\top$, and $\mbf{k} = [k(\btht,\btht_1),\ldots,k(\btht,\btht_n)]^\top$. 

The emulation properties of the GP are driven by the choice of the mean and covariance functions. In this work, we standarize the observations $\mbf{y}_n$ such that they have a mean of zero; that is we set $\mu(\btht) = 0$ in the prior assumption. On the other hand, we use a covariance function from the Matern class~\cite{stein2012interpolation,cressie1999classes,matern2013spatial}, given by 
\begin{equation}
    k(\btht, \btht') = \f{2^{1-\nu}}{\Gamma(\nu)}\left(\f{\sqrt{2\nu} \|\btht-\btht' \|}{\ell} \right)^\nu K_{\nu} \left(\f{\sqrt{2\nu} \|\btht-\btht' \|}{\ell} \right),
\end{equation}
with positive parameters $\nu$ and $\ell$, where $K_{\nu}$ is a modified Bessel function, $\Gamma()$ is the Gamma function and $\| \cdot \|$ denotes the Euclidean distance. The parameter $\nu$ (which we set to $3/2$) controls the differentiability of the sample paths of the GP and is fixed, whereas $\ell$ is a \emph{lengthscale} parameter that controls the rate of change of the sample paths in $\mcl{T}$ and is a hyperparameter. The GP hyperparameter set $\Omega = \{\ell, \sigma_\epsilon^2\}$ is estimated via a maximum likelihood estimation (MLE) procedure frequently followed in fitting GP models~\cite{williams2006gaussian,santner2003design}.

The fact that the posterior distribution of the function given in \eqref{e:posterior} is available in closed-form, the inference of such a model is called \emph{exact} inference. One of the main bottlenecks of the exact inference is that the computation of the posterior mean and variance in \eqref{e:posterior} involves the inversion of the matrix $\mbf{K}$, whose computational cost scales as $\mcl{O}(n^3)$, and hence gets expensive as $n$ increases. This motivates the \emph{approximate} inference technique for GPs, namely, variational GP regression, which trades some accuracy for large gains in computational efficiency in fitting GP models when $n$ is large.

\subsection{Approximate Gaussian process regression}
\label{ss:app_gp}
In addition to the cost of estimating hyperparameters of the exact GP involving $\mcl{O}(n^3)$ floating point operations, the prediction with the GP costs $\mcl{O}(n)$ and $\mcl{O}(n^2)$ operations, respectively, for the posterior mean and variance computation. Sparse GP models~\cite{snelson2005sparse} circumvent this overhead by identifying a subset $m<n$ of the training points, resulting in \emph{reduced} computational costs that scale as $\mcl{O}(m^2n)$, $\mcl{O}(m)$, and $\mcl{O}(m^2)$ for fitting, predicting mean, and predicting variance, respectively. 

The choice of the subset of $m$ inducing points can be treated as another hyperparameter and estimated by maximizing the marginal likelihood, just like in the exact GP model; this results in an extended set of hyperparameters $\{\Omega, \bar{\bTht} \}$, where $\bar{\bTht} \in\mathbb{R}^{m\times d}$ are the \emph{inducing points}.

While sparse GPs bring down the cost of inverting the covariance matrix $\mbf{K}$ and predictions with the GP, the number of hyperparameters increases (due to the addition of the inducing points), thereby making inference computationally more expensive. To overcome this, we use variational inference~\cite{blei2017variational}, which provides a tractable alternative to approximate unknown probability densities in Bayesian models. Below, we briefly provide an overview of sparse GP models and variational inference.

We now present a sparse model that is computationally tractable in terms of inference and prediction with GPs. The sparsity arises because we consider a sparse dataset $\bar{D}$ of size $m<n$ with inducing inputs $\bar{\bTht} = \{\bar{\btht_i},~i=1,\ldots,m\} \subset \mcl{T}$. These inducing inputs can either be a subset of the training inputs or can be randomly sampled from $\mcl{T}$~\cite{snelson2005sparse}. Let $u=f(\bar{\btht})$ be the inducing outputs, which are sampled from the same prior on the true function $f$. In this work, we treat $\bar{\bTht}$ as hyperparameters and estimate them from maximizing the marginal likelihood.

Let the prior on the inducing outputs be given as
\begin{equation}
    p(\mbf{u}|\bar{\btht}) \sim \mcl{N}(0, \mbf{K}_{mm}),
    \label{e:ind_prior}
\end{equation}
where $\mbf{K}_{mm}$ is the matrix of covariances between the inducing inputs. This prior follows from the assumption that the inducing outputs also behave like the latent variable $f$ which has a Gaussian prior. Therefore, $\mbf{u}$ and $\mbf{f}$ have a joint Gaussian distribution~\cite{hensman2015scalable} given by 
\begin{equation}
    \begin{split}
    p(\mbf{f}, \mbf{u}) =& p(\mbf{f|\mbf{u}})p(\mbf{u}) \\
    =& \mcl{N}(\mbf{K}_{nm}\mbf{K}^{-1}_{mm}\mbf{u}, \mbf{K}_{nn} - \mbf{Q}_{nn}) \times \mcl{N}(\mbf{0}, \mbf{K}_{mm}),
    \end{split}
\end{equation}
where $\mbf{Q}_{nn} = \mbf{K}_{nm} \mbf{K}^{-1}_{mm} \mbf{K}_{nm}^\top$. To estimate the extended hyperparameter set $\{\Omega, \bar{\bTht} \}$, we first need to define a marginal likelihood. In this case, the marginal likelihood is marginalized over $\mbf{f}$ and $\mbf{u}$, that is,
\begin{equation}
    \begin{split}
    p(\mbf{y}|\bTht, \Omega, \bar{\bTht}) =& \int p(\mbf{y}|\mbf{u}) p(\mbf{u}) d\mbf{u} \\
    \end{split}
    \label{e:svgp_likelihood}
\end{equation}
Note that the density $p(\mbf{y}|\mbf{u})$ still involves inverting the matrix of size $n\times n$ and hence is expensive to compute. Therefore, this density is approximated using the evidence lower bound (ELBO)~\cite{blei2017variational} as
\begin{equation}
    p(\mbf{y}|\mbf{u}) \geq \mbb{E}_{p(\mbf{f}|\mbf{u})} \left[ p(\mbf{y}|\mbf{f})\right].
    \label{e:elbo}
\end{equation}

Substituting \eqref{e:elbo} in \eqref{e:svgp_likelihood}, the lower bound on the marginal log likelihood can be approximated as~\cite{hensman2015scalable}
\begin{equation}
    p(\mbf{y}|\bTht, \Omega, \bar{\bTht}) \geq \T{log} \mcl{N} (\mbf{0}, \mbf{K}_{nm} \mbf{K}^{-1}_{mm} \mbf{K}_{nm}^\top + \sigma^2_\epsilon \mbf{I}) - \f{1}{2} \T{tr}(\mbf{K}_{nn} - \mbf{Q}_{nn}).
    \label{e:svgp_variational_likelihood}
\end{equation}

\Cref{e:svgp_variational_likelihood} can be maximized with respect to $\bar{\bTht}$ and $\Omega$ to estimate the hyperparameters, where the bound in \eqref{e:svgp_variational_likelihood} costs $\mcl{O}(nm^2)$ for computation.

\subsection{Active learning for Gaussian process regression}
\label{ss:al_gp}
Whereas variational GPs provide an approximation to exact GP regression, another approach to improving tractability of exact GP models is active data selection ~\cite{cohn1995active,cohn1996active}. Specifically, given a training data set $\mcl{D}_n$, the data set is adaptively augmented as
\begin{equation}
\begin{split}
    \mcl{D}_{n+i} & := \mcl{D}_{n+i-1} \bigcup \{(\btht_{n+i}, y_{n+i})\}, ~i=1,\ldots,m
\end{split}
\end{equation}

where $m$ is the number of adaptive model building steps. Furthermore, each adaptive step can select a batch of $q$ training points jointly; when $q=1$ we call the approach \emph{sequential}  active learning and when $q>1$, \emph{batch-sequential} active learning. At the end of the active learning process, the GP is trained with a total of $n + mq$ training points. The main objective of active learning is to choose points judiciously such that they are \emph{optimal} in the sense of improving model fit.

In this work, we choose points that are optimal in reducing the overall uncertainty about the GP model. That is, we select points such that
\[ \btht_{n+1} = \T{arg} \max_{\btht' \in  \mcl{T}}~\int_\mcl{T} -\sigma_{n+1}^2(\btht') ~d\btht' \]
where $\sigma_{n+1}^2(\btht)$ is the posterior variance of the GP, having observed the $(n+1)$th point, and we introduce the negative sign to solve a maximization problem. Essentially, we treat the integrated posterior variance as a measure of uncertainty over our domain $\mcl{T}$ and seek to choose training data that are optimal in minimizing this uncertainty.

A fundamental issue with the above equation is that the posterior variance $\sigma_{n+1}^2(\btht)$ is unknown until we actually commit to $\btht_{n+1}$ and choose a training point there. To circumvent, we simulate the choice of $\btht_{n+1}$ via the GP trained with data $\mcl{D}_{n+1}$, and choose $\btht_{n+1}$ as the point that reduces the \emph{expected} uncertainty:
\begin{equation}
    \begin{split}
    \btht_{n+1} =& \T{arg} \max_{\btht' \in  \mcl{T}}~ - \int_{\mcl{Y}} \int_\mcl{T} \sigma_n^2(\btht) | \mcl{D}_n \cup \{\btht', y(\btht')\}~d\btht dy \\
    = & \T{arg} \max_{\btht' \in  \mcl{T}}~- \mbb{E}_{y\sim Y_n} \left[ \int_\mcl{T} \sigma_n^2(\btht) | \mcl{D}_n \cup \{\btht', y(\btht')\}~d\btht \right].
    \end{split}
    \label{e:imse}
\end{equation}

Essentially, \eqref{e:imse} seeks to find the point in $\mcl{T}$ that likely leads to the least overall uncertainty, if the corresponding training point was chosen and the model updated. 

Similarly, the batch-sequential active design, to select $\Tht = \{\btht_1,\ldots,\btht_q\}$, is performed by choosing points as
\begin{equation*}
    \btht_{n+1:q} =  \T{arg} \max_{\Tht'}~ - \mbb{E}_{y\sim Y_n} \left[ \int_\mcl{T} \sigma_n^2(\btht) | \mcl{D}_n \cup \{\Tht', y(\Tht')\}~d\Tht \right].
\end{equation*}

The choice of $q>1$ particularly has advantages when the training data are generated by running expensive computer simulations, which can be evaluated synchronously in parallel. In the cases such as the present work, where we seek to actively select training data from an existing set, batch-sequential selection results in fewer hyperparameter training steps, which in the case of exact GPs scales as $\mcl{O}(n^3)$. However, in terms of the improvement in model fit, it is not obvious what choice of $q$ is the best. Therefore, we perform a simple sensitivity study to investigate the effect of the choice of $q$ on the model fit.

Finally, to assess model fit, we evaluate the GP posterior mean on a hold-out test set $\{\btht_i, f(\btht_i) \},~i=1,\ldots,n_\T{test}$, and compute the log root mean squared error (RMSE), defined as
\begin{equation}
    \log(RMSE) = \log \left\{ \f{1}{n_\T{test}} \sum_{i=1}^{n_\T{test}} [\mu(\btht_i) - f(\btht)_i]^2 \right\}^{1/2}.
\end{equation}
We note that the log(RMSE) is computed at each step of the active learning process with the GP model trained with the training data selected up to the previous step. Furthermore, the log(RMSE) is independently computed for each of the latent space outputs using the corresponding GP model.

\section{Results}
\label{s:results}
We now demonstrate our proposed data-driven machine learning approaches toward predicting the wake velocity field, using data generated from a scanning wind LiDAR. We begin by first discussing the latent space reconstruction accuracy of the data, purely from the convolutional autoencoder. The reconstruction results provide a visual estimate of the trade-off between compression due to the autoencoder and loss/retention of information. With the latent space representation available from the autoencoder compression, we then proceed to learn the mapping between the inputs (operating conditions) and the latent space, via the machine learning models. Finally, the predicted wake fields are "decoded" via the decoder, as shown in \cref{fig:method}.

\subsection{Compression accuracy}
\label{ss:comp_accuracy}
\begin{figure}
    \centering
    \includegraphics[width=\textwidth]{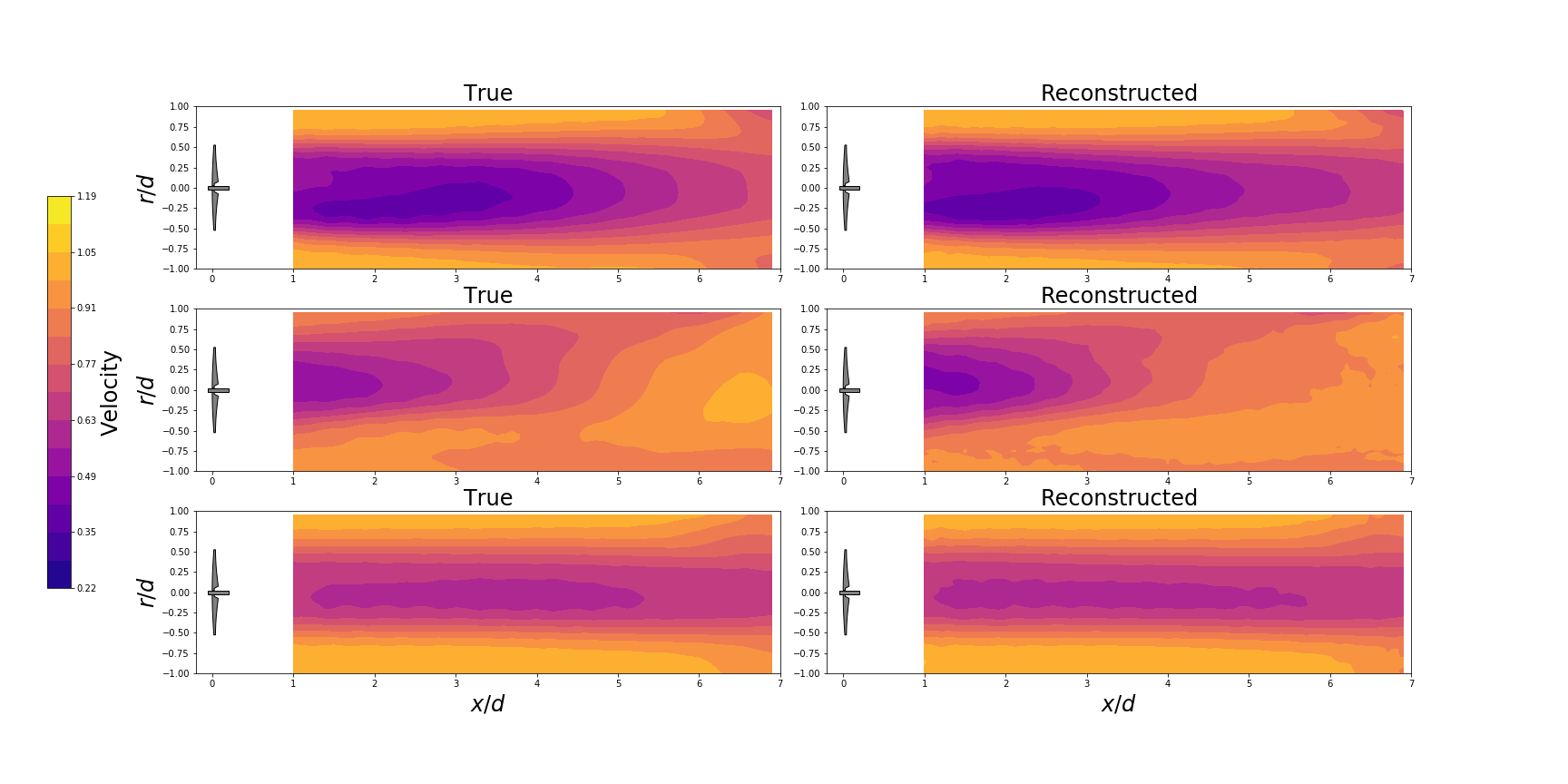}
    \caption{The compressive effectiveness of the convolutional autoencoder on the LiDAR data set for 3 test examples. The left column shows the true behavior of the wake for a set of testing data (unseen during model fit) and the right column shows reconstruction from a 4-dimensional latent space by the autoencoder.}
    \label{fig:recon_1}
\end{figure}

We first examine the ability of the convolutional autoencoder to effectively compress the LiDAR observations to a suitable latent space. \cref{fig:recon_1} shows the ability of the autoencoder to reconstruct observed data, with only four latent dimensions, through its bottleneck neural architecture. The figures demonstrate that despite the drastic dimensionality reduction (i.e., 2501 to 4), the reconstruction accuracy has not been compromised significantly. However, the larger goal here is that we want to be able to generalize to a similar reconstruction error everywhere in the parameter space. For this, we introduce the results of the methods to obtain parameter-output maps, which we present next.



\subsection{Parameteric reconstruction}
\label{ss:param_reconstruction}

The latent space $\tilde{\x} \in\mathbb{R}^{k}$ provides a very concise encoding of the high-dimensional velocity field since $k \ll d$. Therefore, we learn the function $f: \mbf{\theta} \in\mathbb{R}^p \rightarrow \tilde{\x} \in\mathbb{R}^{k}$. We learn $f$ via MLP and GP regression (with exact inference). Furthermore, we also use VGP regression (approximate inference) to improve the tractability of GP models for large datasets. Finally, we also show the performance of choosing training data via active learning for the exact GP.

\begin{figure}
    \centering
    \mbox{
    \subfigure[Dimension 1]{\includegraphics[width=0.49\textwidth]{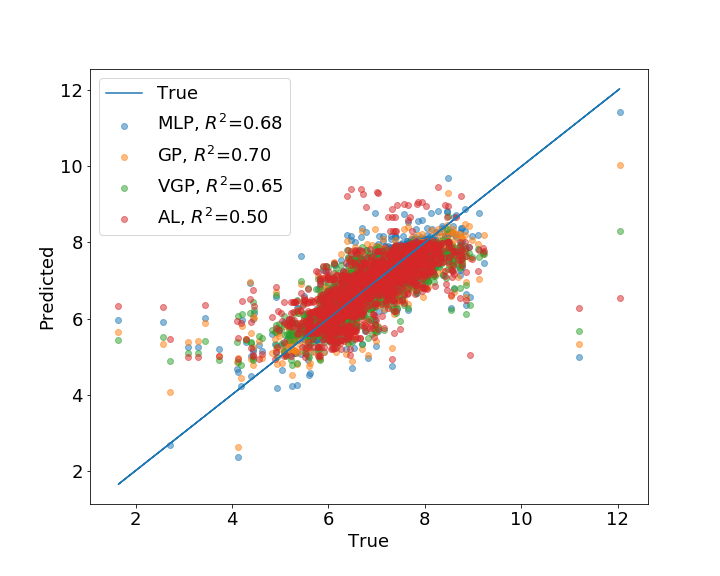}}
    \subfigure[Dimension 2]{\includegraphics[width=0.49\textwidth]{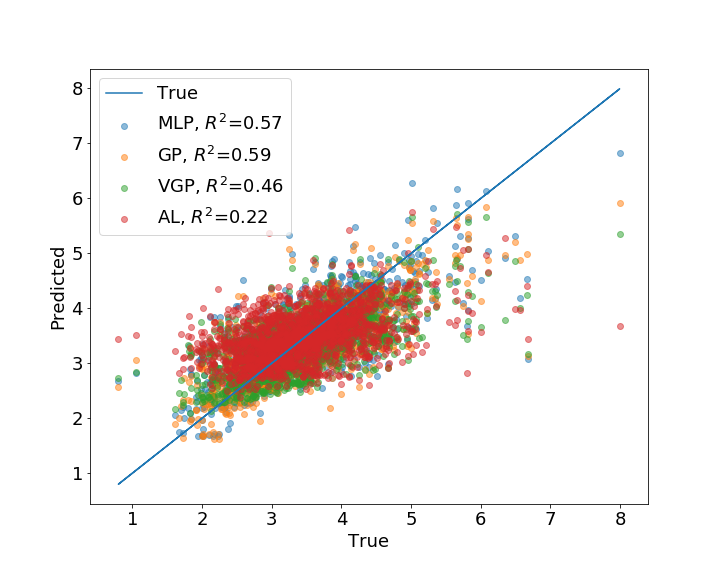}}
    } \\
    \mbox{
    \subfigure[Dimension 3]{\includegraphics[width=0.49\textwidth]{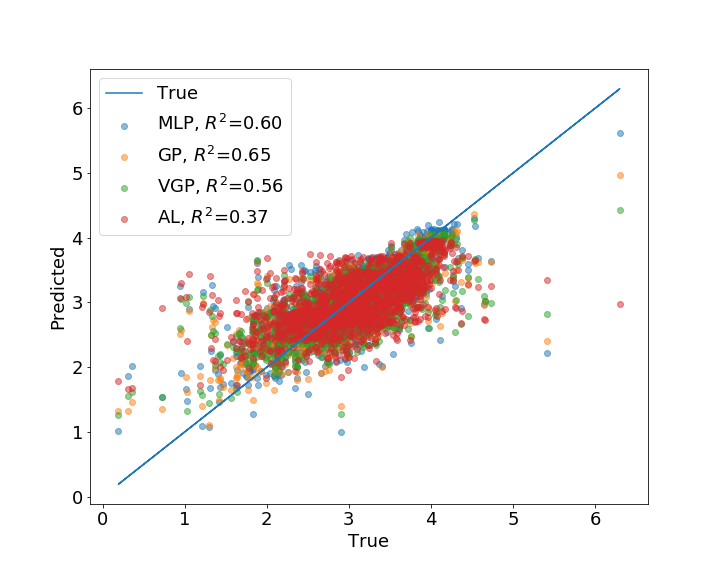}}
    \subfigure[Dimension 4]{\includegraphics[width=0.49\textwidth]{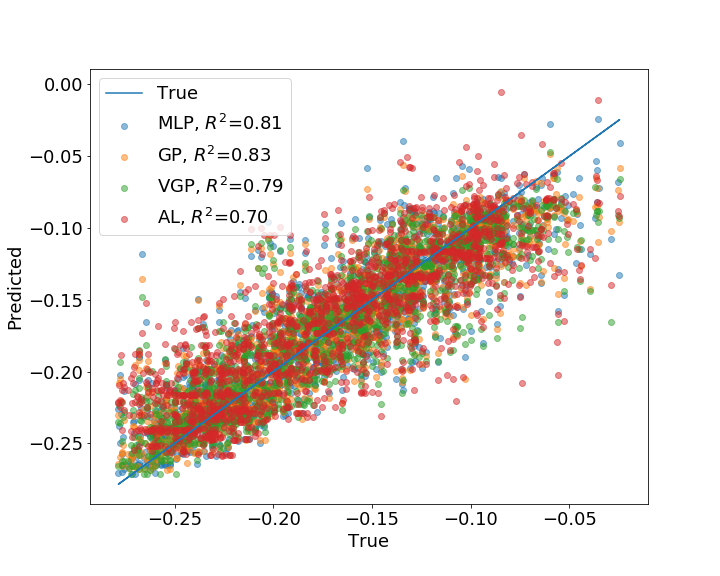}}
    }
    \caption{Fitting a parameteric map between MET data and the latent space representation extracted by the convolutional autoencoder on the LiDAR snapshots. The different colors indicate different latent space parameteric mapping techniques. Similar results are obtained across different methods. }
    \label{fig:param_nn_gp}
\end{figure}

The original LiDAR dataset has a dimensionality $d=2501$ which is reduced to $k = 4$, via the convolutional autoencoder. \cref{fig:param_nn_gp} shows the actual-vs-predicted plot of each of the four latent space dimensions predicted via all of the three machine learning models: MLP, GP, and VGP; note that we also show results of fitting GP with active learning used for training data selection. Firstly, the plots show that the predictive accuracy for all the machine learning models are somewhat similar. Secondly, there are outliers in the dataset---as in, for example, the lower-left and upper-right corners of subfigure $(a)$, where the prediction accuracy is poor. We attribute this primarily to the fact that the LiDAR measurements are corrupted by noise, whose structure is unknown and not captured by the models. Even though the GP models do resolve the noise in the dataset, our model makes simplifying assumptions such as independent and identically distributed noise, which might not necessarily provide a realistic model of the noise (although they are relatively computationally more tractable). Furthermore, the raw measurements have missing elements, which are imputed via a local interpolation, which---although inevitable---is expected to bias the dataset. Given these characteristics of our real-world dataset, including more sophistication, such as heteroscedastic noise variance, could potentially overfit the data; we reserve those approaches for future work.

\begin{figure}
    \centering
    \mbox{
    \subfigure[Example 1]{\includegraphics[width=\textwidth]{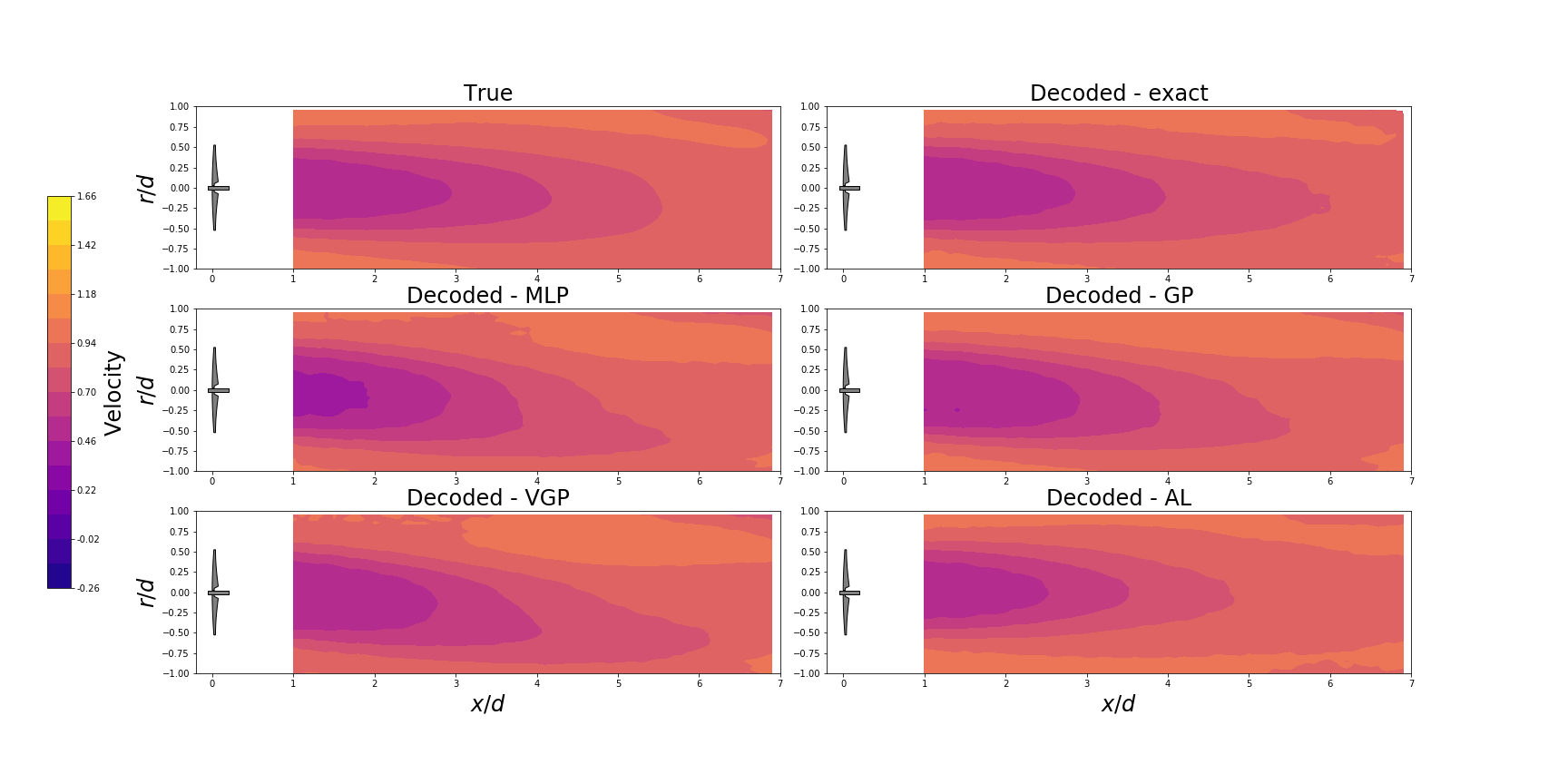}}
    }\\
    \mbox{
    \subfigure[Example 2]{\includegraphics[width=\textwidth]{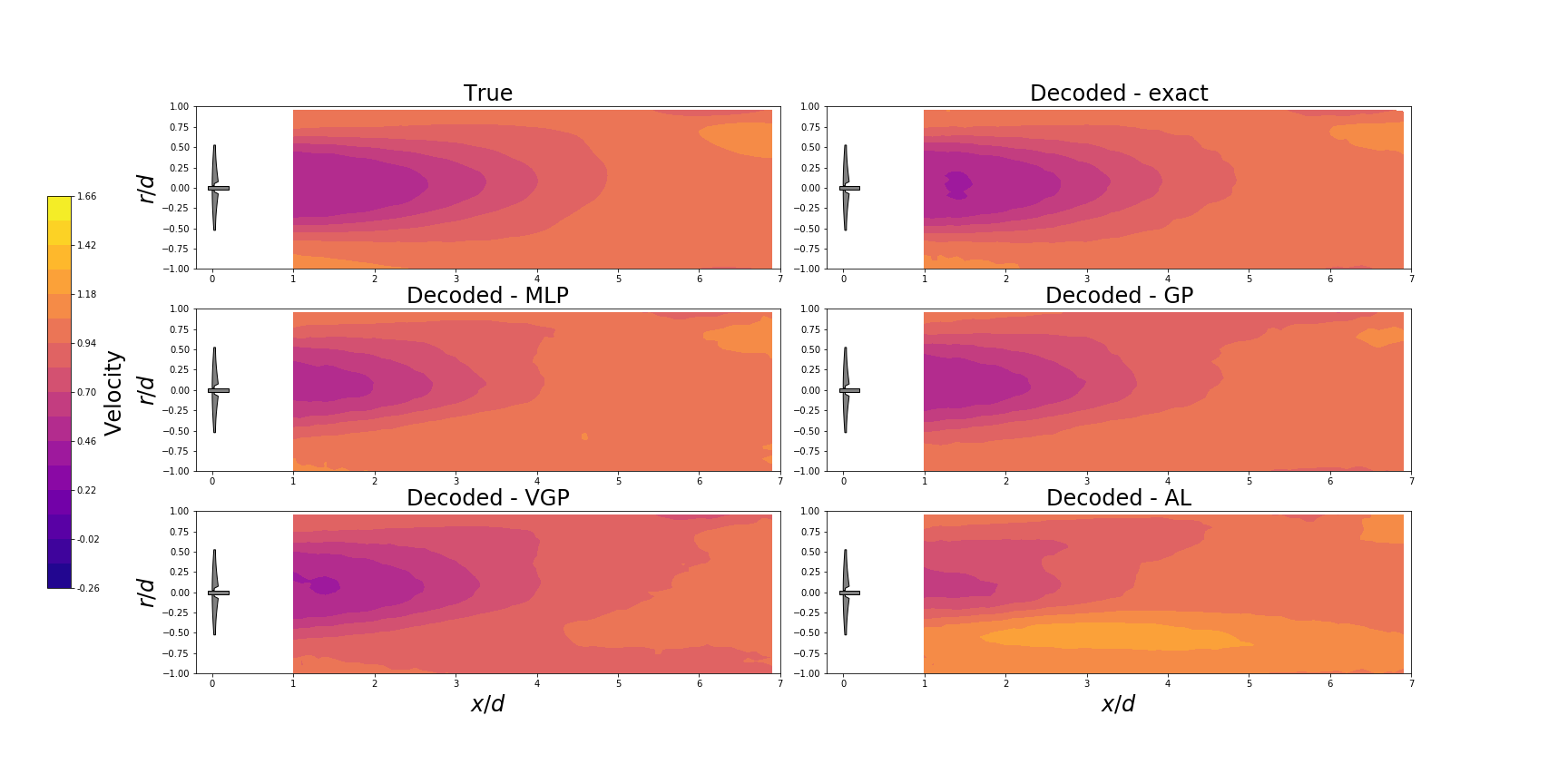}}
    }
    \caption{Two examples of parameteric reconstruction ability where information from the MET data is used to obtain a latent-space representation for a test data point. Following this, the decoder of the autoencoder is used to reconstruct in physical space. We show results for the true test snapshots, their exact reconstructions using only the autoencoder, and various parameteric predictions in latent space followed by use of the decoder.}
    \label{fig:param_rec_1}
\end{figure}

We show the predicted flow field with our machine learning models, for two unseen parameters in \cref{fig:param_rec_1}. In addition to the true flow-field obtained from experimental observations, and the \emph{best possible} reconstruction---via \emph{autoencoding} the true flow-field, the prediction accuracy is more-or-less uniform across the various models, and visually presents a very close match to the true flow field. Note that, the machine learning models at best can only emulate the reconstruction decoded via the autoencoder and hence comparison against the \emph{Decoded-exact} plot is most appropriate. 

\begin{figure}
    \centering
    \includegraphics[width=\textwidth]{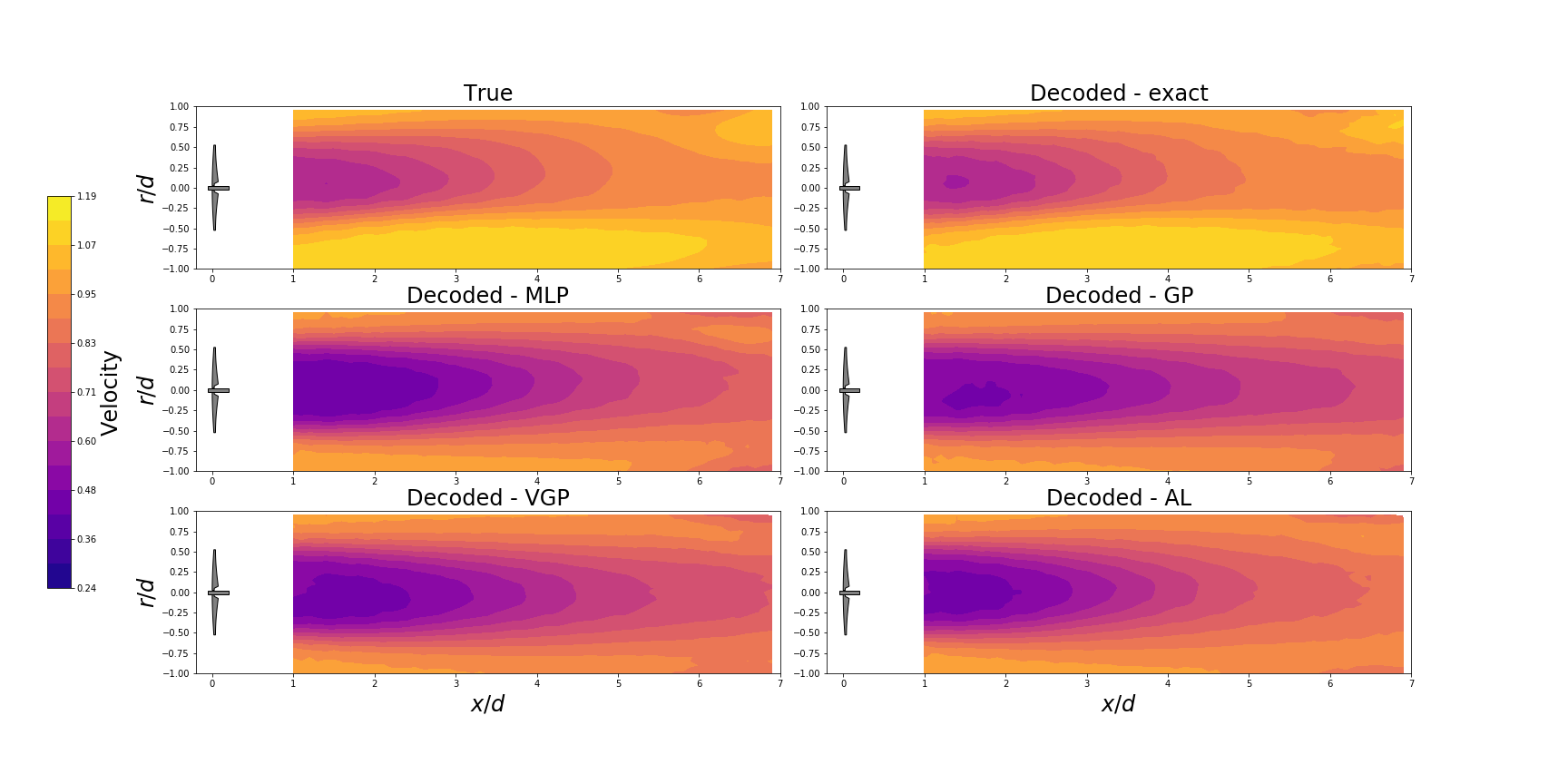}
    \caption{Potential issue with an outlier - showing limited parameteric map expressiveness across various latent space interpolation methods. Note that outliers were limited to around 10\% of the dataset.}
    \label{fig:param_rec_2}
\end{figure}

In \cref{fig:param_rec_2}, we also show a \emph{worst-case} prediction via our machine learning models. These plots correspond to the outliers identified in \cref{fig:param_nn_gp}. Overall, we noticed that total number of such outliers fall roughly within 10\% of the overall dataset. For the specific wake measurement shown in \cref{fig:param_rec_2}, we notice that an anomalous speed-up is observed on the side of the wake for negative values of the radial position. This flow feature can be either due to flow interaction with side wind turbines or to large coherent flow structures typically present in the atmospheric boundary layer. Even though this kind of wake realizations are realistic, their occurrence can be relatively low and, thus, not captured from the training dataset.

\begin{figure}[htb!]
    \centering
    \includegraphics[width=1\textwidth]{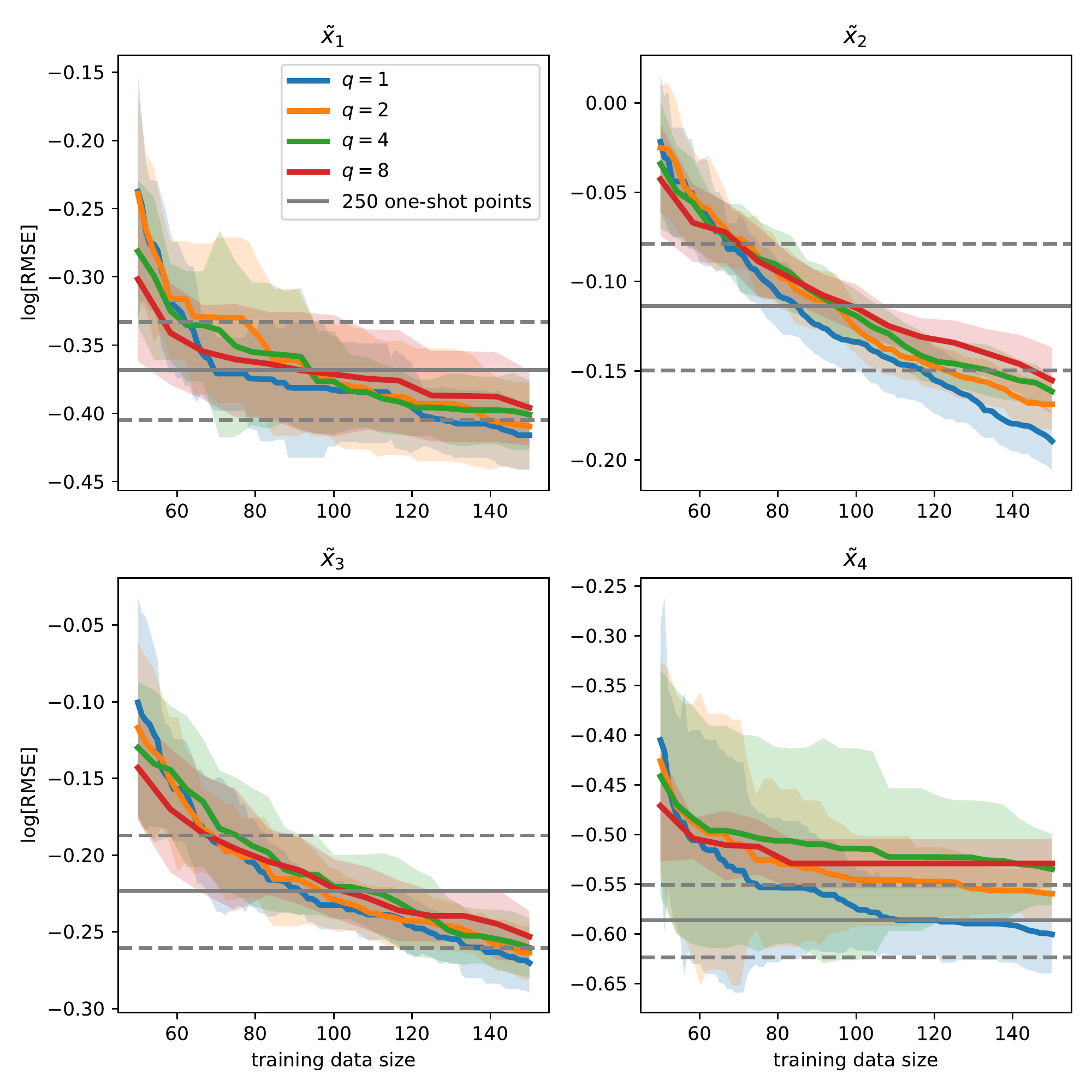}
    \caption{One-shot vs sequential GP fit. Horizontal lines are the RMSE with a one-shot GP fit with 500 uniformly random points; solid and dashed lines are the mean and $\pm 1 \sigma$, respectively. Sequential fit outperforms one-shot fit despite fewer training data.}
    \label{fig:rmse}
\end{figure}

We have used the sequential GP model fitting via AL is another approach to improve the tractability of (exact) GP regression for large datasets. Essentially, we choose the most \emph{relevant} subset of the training data set, instead of using the entire data set or randomly sampling from it. \cref{fig:rmse} shows the log RMSE of the GP---based on a held out test data set of 1781 points---with sequential increments in the training data. Recall that the points are chosen sequentially to reduce the average uncertainty about the GP in the entire input domain $\mcl{X}$, and \cref{fig:rmse} shows the resulting prediction error (log RMSE) variation with training data set size. We start the GP fit with a randomly selected 50 points (from the full data set of 5000 points) and sequentially add 100 more points. We also show the impact of the choice of the number of points selected ($q$) at each step by varying it between $q=1$ to $q=8$. We repeat this exercise 20 times for independently chosen random starting points and plot the mean and $\pm 1$ standard deviation. Finally, we also show the prediction error due to selecting 250 points randomly in \emph{one shot} (i.e., no sequential point selection). The results show that sequential point selection results in a smaller prediction error, regardless of the choice of $q$, for the first 3 latent space dimensions. For the fourth latent space dimension ($\tilde{\x}_4$), the $q=1$ still outperforms the one-shot selection. It is worth noting that the one-shot selection of points still has 100 points more than what was supplied to the AL GP. The reason for the AL GP outperforming the one-shot GP is because training points are more judiciously chosen--in this case, they are chosen specifically to minimize the uncertainty in the GP about its own prediction. Furthermore, it can be shown~\cite{santner2003design} that the average uncertainty in the GP is equivalent to the mean-squared prediction error (MSPE) of the posterior mean of the GP, and therefore in effect the AL chooses points to minimize the overall prediction error. 

\begin{figure}[htb!]
    \centering
    \includegraphics[width=1\textwidth]{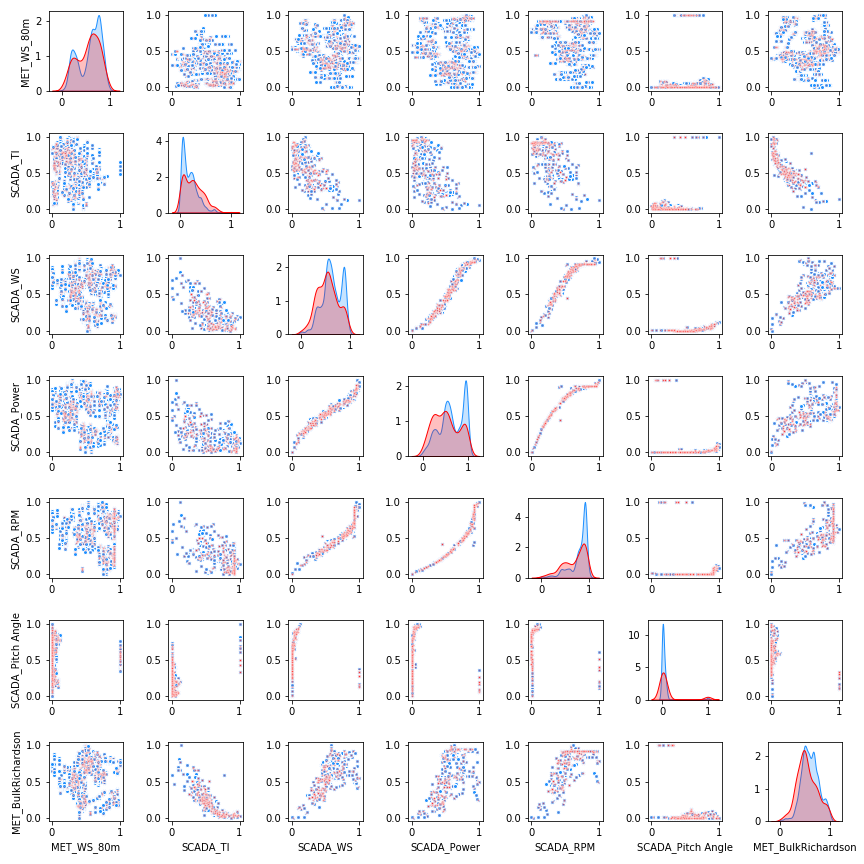}
    \caption{Selected points via active learning. Blue represents the full training data set (5000 points) and red represents the points selected via active learning (150 points).}
    \label{fig:scatter}
\end{figure}

The spatial distribution of points is visualized in \cref{fig:scatter}, which is a scatterplot matrix of all possible combinations of the inputs listed in \cref{tab:input_parameters}. In this figure, the blue symbols indicate the full training data set (5000 points) and the red symbols are the points selected via AL. Notice that the red points show a better spread and hence coverage of the design space compared to the full data set. This is further emphasized by the kernel density plots shown along the diagonal of the scatterplot matrix, where the AL shows a larger variance which is indicative of the fact that points are more spread out. Overall, we see that the choosing points via AL another way toward building a more tractable GP model with large data sets, without compromising on the predictive accuracy.

\begin{figure}
    \centering
    \mbox{
    \subfigure[Neural network]{\includegraphics[width=0.48\textwidth]{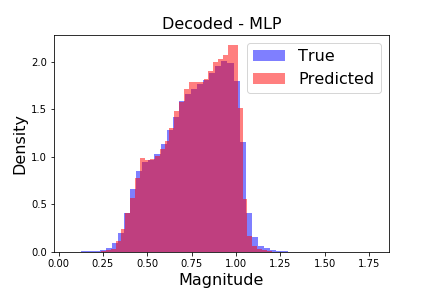}}
    \subfigure[Gaussian process]{\includegraphics[width=0.48\textwidth]{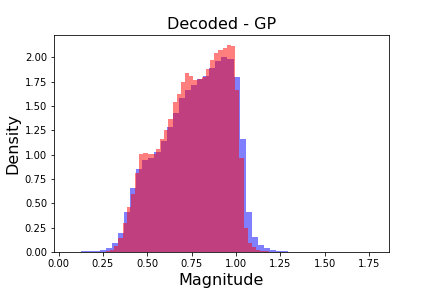}}
    } \\
    \mbox{
    \subfigure[Variational Gaussian Process]{\includegraphics[width=0.48\textwidth]{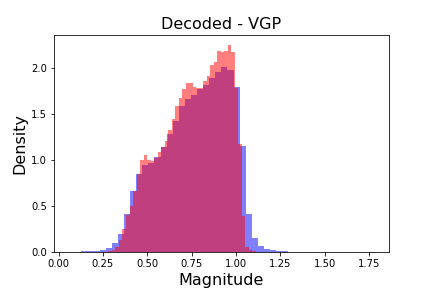}}
    \subfigure[Active Learning]{\includegraphics[width=0.48\textwidth]{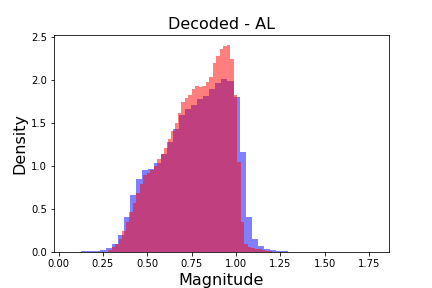}}
    }
    \caption{Density plots of data-driven predictions in latent space followed by reconstruction through the decoder for (a) a fully connected neural network (b) a Gaussian process (c) a sparse variational Gaussian process and (d) a reconstruction using a Gaussian process trained by active learning.}
    \label{fig:bins}
\end{figure}

\begin{figure}
    \centering
    \mbox{
    \subfigure[Sorted Error]{\includegraphics[width=0.48\textwidth]{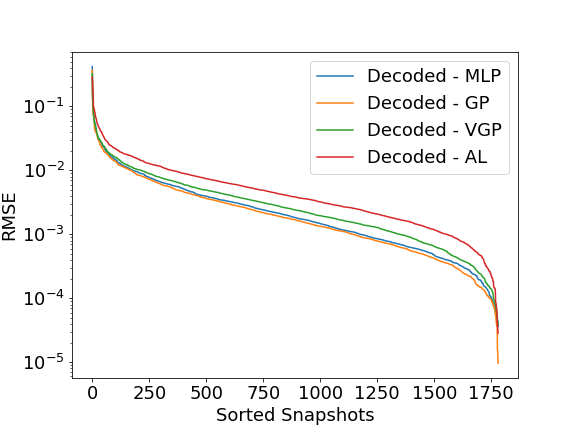}}
    \subfigure[Kernel density estimate of MSE]{\includegraphics[width=0.48\textwidth]{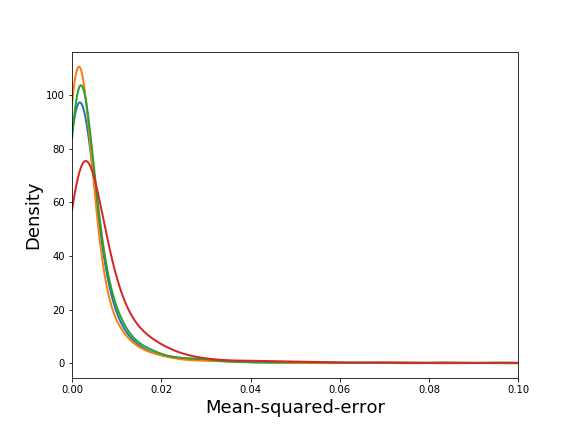}}
    }
    \caption{The sorted root-mean-squared error on the testing data set after parameteric latent space prediction and reconstruction in physical space (left). Kernel density estimate of the mean-squared errors for different parametric maps (right). The RMSEs are competitive across a large portion of the test data set with certain outliers (approximately 200 data points).}
    \label{fig:outliers}
\end{figure}

\cref{fig:bins} shows a probability density plot of the prediction accuracy for all the machine learning models we employed in this work. Note that the densities have similar shapes indicating that the overall predictive accuracy for all the used models is somewhat similar, as mentioned previously. It is also worth noting that the AL GP still has the lowest accuracy amongst all the three GP approaches presented; this can be appreciated by considering the $R^2$ value in \cref{fig:param_nn_gp} and/or the sorted error and kernel density plot of the error shown in \cref{fig:outliers}. Our hypothesis for this behavior is that these plots show just one realization of the GP (as opposed to \cref{fig:rmse} where we show the average of 20 independent realizations). In general, the predictive accuracy of the GP is sensitive to the choice of starting points, but \cref{fig:rmse} proves that on average, the AL GP outperforms a GP with one-shot point selection. Despite this fact, we still show just one realization of the AL GP in \cref{fig:param_nn_gp} and \cref{fig:outliers} (despite showing a relatively poor predictive accuracy) because this is more representative of a realistic scenario.

\section{Conclusions}
\label{s:conclusions}
Accurate predictions of wind-turbine wake flows are crucial to estimate power efficiency of wind farms, power losses due to wake interactions, and optimal design of wind farm layout. High-fidelity large eddy simulations, which can resolve the turbulent flow fields with high accuracy, are prohibitively expensive from the computational-cost standpoint to tackle the foregoing wind energy problems. Scanning Doppler wind LiDARs provide detailed measurements of the wake flow field generated from utility-scale wind turbines; however, experimental data can be incomplete and noisy. 

In this work, we develop data-driven machine learning models for prediction of wind turbine wakes by leveraging wind LiDAR measurements. Specifically, we develop several data-driven models from high-dimensional LiDAR data and analyze their comparative performance in terms of parametric prediction. In this regard, we first use deep neural networks to achieve a drastically compressed ($2501 \rightarrow 4$) latent-space representation of the high-dimensional data. Then, we use multilayered perceptrons and Gaussian processes to learn the input parameter-latent-space map. Results show that the predictive capability of all the machine learning models are somewhat similar. However, GP regression with exact inference resulted in the least RMSE (best prediction). Furthermore, to address the well-known tractability issues with exact-inference Gaussian processes, we also propose variational sparse Gaussian processes as well as active learned Gaussian processes. These two approaches, while resulting in a significant saving in the computation necessary for model inference, are observed to make only a marginal trade in the accuracy. Overall, once trained, all of the machine learning model takes only $O(1)$ second of wall-clock time to evaluate.

This work brings together state-of-the-art machine learning, to develop data-driven models of real-world data pertinent to wind energy. We show that robust and accurate models can be developed despite the data being noisy and incomplete. With the developed predictive models, sensitivity of the wake flow field to the input parameters can be analyzed in real-time, compared to the unrealistic cpu-hours necessary for high-fidelity simulations. In the future, we hope to streamline the whole framework for data-driven wake modeling via automated neural architecture search and improved uncertainty quantification (e.g., using heteroscedastic noise models for Gaussian processes). Furthermore, developing hybrid probabilistic machine learning models such as deep GP models, which combine multilayered perceptrons and Gaussian processes, is another focus of future work. 

\clearpage
\section*{Acknowledgement}
This material is partially based upon work supported by the U.S. Department of Energy (DOE), Office of Science, Office of Advanced Scientific Computing Research, under Contract DE-AC02-06CH11357. This research was funded in part and used resources of the Argonne Leadership Computing Facility, which is a DOE Office of Science User Facility supported under Contract DE-AC02-06CH11357. SAR acknowledges the support by Laboratory Directed Research and Development (LDRD) funding from Argonne National Laboratory, provided by the Director, Office of Science, of the U.S. Department of Energy under contract DE-AC02-06CH11357. This research has been partially funded by a grant from the National Science Foundation CBET Fluid Dynamics, award number 1705837. Pattern Energy Group is acknowledged to provide access to the wind farm for the LiDAR experiment and wind farm data. 

\begin{mdframed}
    The submitted manuscript has been created by UChicago Argonne, LLC, Operator of Argonne National Laboratory ("Argonne”). Argonne, a U.S. Department of Energy Office of Science laboratory, is operated under Contract No. DE-AC02-06CH11357. The U.S. Government retains for itself, and others acting on its behalf, a paid-up nonexclusive, irrevocable worldwide license in said article to reproduce, prepare derivative works, distribute copies to the public, and perform publicly and display publicly, by or on behalf of the Government. The Department of Energy will provide public access to these results of federally sponsored research in accordance with the DOE Public Access Plan (http://energy.gov/downloads/doe-public-access-plan).
\end{mdframed}

\bibliographystyle{unsrt}  
\bibliography{references}  

\end{document}